# Three New Methods for Evaluating Reference Resolution

**Andrei Popescu-Belis & Isabelle Robba**
Language and Cognition Group
LIMSI – CNRS
BP 133, 91403 ORSAY Cedex, FRANCE
{popescu, robba}@limsi.fr

**Abstract**

Reference resolution on extended texts (several thousand references) cannot be evaluated manually. An evaluation algorithm has been proposed for the MUC tests, using equivalence classes for the coreference relation. However, we show here that this algorithm is too indulgent, yielding good scores even for poor resolution strategies. We elaborate on the same formalism to propose two new evaluation algorithms, comparing them first with the MUC algorithm and giving then results on a variety of examples. A third algorithm using only distributional comparison of equivalence classes is finally described; it assesses the relative importance of the recall vs. precision errors.

## Reference Resolution and its Evaluation

In the field of natural language processing, reference resolution, i.e., linking phrases that designate the same entity, has been a long-standing cornerstone. Approaches of the problem differ basically according to their grasp of real world: if the system possesses a world-model, then it resolves reference by linking referring expressions of the text to representations of real entities. Otherwise, without a world-model, the system can only group together expressions pointing to the same real entity. Evaluation oscillates also between these two approaches, but a general method should be able to apply to both situations.

In this paper, we formalize the main notions underlying evaluation, and comment on an existing scoring method, used in the Message Undertstanding Conference tests (MUC-6). After showing some shortcomings of this method, we define a new principle- based evaluation algorithm and compare it with the former from both theoretical and experimental points of view. An extension of this algorithm is then discussed; a third algorithm is finally introduced on pure statistical grounds.

## Reference or Coreference ?

**Definitions** It is generally admitted that a discourse contains *referring expressions (REs)* which designate discourse referents. If a program can emulate human comprehension of the discourse, it builds and maintains a set of *mental representations (MRs)* of the referents, and attaches each RE to the proper MR; this is reference resolution. A program may build only coreference links between REs that refer to the same entity, for instance anaphoric links between a pronoun and a noun phrase; this is co-reference resolution.

The two approaches share an important common feature: as coreference is an *equivalence relation*, both handle in fact *equivalence classes of REs*, the first one explicitly, the second one after construction of the transitive closure of all coreferences. These classes constitute the core of Vilain's et al. (1995) theoretical model for evaluation, and we rely on them as well. Our method applies thus to systems which produce equivalence classes as well as to those producing coreference relations which enable us to build equivalence classes.

**Examples of Systems** Reference solving systems can be classified according to the previous view. Some systems are concerned solely with anaphoric coreference (pronouns), e.g., algorithms by Hobbs (1978) or the one by Lappin & Leass (1994). Some others handle general coreference, between pairs of REs, as requested by the MUC-6 coreference task (Sundheim & Grishman 1995) – for instance the UMass system (Fisher et al. 1995). More elaborated discourse understanding systems build representations of discourse referents, and attempt to attach REs not to other REs but to referent representations, for instance Luperfoy's (1992) system. We are currently developing a system using this kind of approach (cf. Popescu-Belis & Robba 1997, 1998 and Popescu-Belis 1998).

## Evaluation

**Requirements** It is uneasy to define precisely the kind of information one expects from an evaluation algorithm. In general, one needs to compare the program's output or *response* to the correct answer or *key*[1]; here, evaluation without a *key* seems meaningless, because it would need supplementary knowledge that would be in fact more useful inside the program, to find a better solution. So, the intuitive idea that evaluation is a measure of how good a program is can be reformulated as: evaluation is a measure of the *difference between the key and the response*.

The simplest test is identity: is the response the same as the key ? Obviously, more is needed: how close to the key is the response ? In very general terms, an evaluation method should return a numeric value (number or tuple) which can be compared to a reference value for the "perfect response", and possibly to other values for the "worst response". But there is no such formal notion of a "fairly good response". A good measure is one which conforms to human intuition of discourse understanding. The MUC-6 method clearly meets the previous requirements; so, the only reason to propose and justify new methods is to give more sensible answers in situations where the MUC-6 method's results seem counterintuitive. This is the main goal of this paper.

---

[1] Even if there isn't always complete agreement on what the key is, as shown by Hirschman et al. (1998).

**Basic concepts** Evaluation measures for elementary discourse understanding have generally been inspired by those used in information retrieval (Salton 83). The system's response is decomposed in three subsets: correct answers, wrong answers and missing answers. The proportional size of the second set gives a *precision error*, while the third set gives a *recall error*. For (co)reference resolution, the information to be found are the coreference links; they define also unique equivalence classes. Evaluation has to define and compute a *recall success* (proportion of retrieved links) and a *precision success* (proportion of correct links). The difficulties lie in computing the relevant number of links, and finding out the proportion. Our approach proposes different counts from the MUC-6 ones, while remaining close to the confirmed recall/precision paradigm.

One has to point out that such a measure returns two values, which already makes comparisons difficult: if a system has better recall but less precision than another, which one is better ? A solution is to combine the two values in one (average, or f-measure[2]) but this problem is in fact part of a more general but unsolvable one: if evaluation produces a relevant answer (with many parameters) than two answers cannot easily be compared - but if it produces only one value, then it isn't enough relevant.

## Use of Evaluation

It is of course obvious that evaluation allows comparison between accomplished systems, and enables developers to improve their programs. Two applications are worth mentioning: automatic evaluation can be used to tune automatically some parameters of reference solvers, as shown in (Popescu-Belis 1998). This is an optimization problem, with a potential field given by the system's scores, depending on the parameters' values. Parameters are modified one by one using a gradient ascent method, in order to improve the scores.

More discrete tuning is also possible: evaluation can assess the proper contribution of each selection rule to the overall success (Popescu-Belis & Robba 1998). So, coreference rules aren't only proposed on a theoretical or empirical base, they can be also validated statistically on long and unrestricted texts.

## The Theoretical Frame

In order to describe clearly the new scoring methods some formal definitions are necessary. These will enable us also to prove some interesting properties related to the requirements above, and to compare our method against the one proposed by Mark Vilain et al. (1995) and used in the MUC-6 and MUC-7 evaluation rounds (Sundheim & Grishman 1995). Vilain's et al. description avoids theoretical developments for the sake of simplicity, but full understanding requires some theory. The frame we propose here is fully compatible with Vilain's et al. description.

### Definitions and Notations

Reference resolution has for input a set $R$ of referring expressions (REs). In general, key and response use different sets, but it makes little difference, for the algorithms shown here, to consider that $R$ is the same[3].

**Partitions** The key is the correct grouping of the REs into equivalence classes (for the coreference relation) or MRs. We note these classes $K_1, K_2, ..., K_n$; they form a partition of $R$ noted $P_K$ because they are always disjoint and they cover $R$. Let $P_K = \{K_1, K_2, ..., K_n\}$[4].

The response is another partition of R, $P_R = \{R_1, R_2, ..., R_n\}$, because the $R_i$ are disjoint and they cover $R$, considering that non-solved REs form singleton MRs.

Comparison between the key and the response becomes in this view a comparison between the two partitions $P_K$ and $P_R$; unfortunately, no mathematical distance between partitions of a set (here $R$) has any relevance for our problem.

**Projections** When the response matches perfectly the key, for each key equivalence class $K_i$ there exists a unique response equivalence class $R_j = K_i$. In "almost perfect" responses, one could try to find for each $K_i$ an $R_j$ so that $K_i \approx R_j$, but in general this is not possible. Instead, it is useful to find out all the $R_j$ containing elements of a given $K_i$. To describe the scoring methods, we define first the projection of $K \in P_K$ on $P_R$ as the set of all non-empty intersections of K with elements of $P_R$ (response MRs):

$$\pi(K) = \{A \mid A = K \cap R_i \text{ with } R_i \in P_R\}.$$

Then, the extended projection $\pi^*$ gives the response classes (MRs) containing the projections $\pi(K)$:

$$\pi^*(K) = \{R_j \mid R_j \in P_R \text{ and } R_j \cap K \neq \emptyset\}.$$

Some elementary properties follow immediately, such as :

$$1 \leq |\pi(K)| \leq |K|^5.$$

Indeed, K has at least one projection and can be split at most in |K| singletons (we use the term projection alternatively for $\pi(K)$ or one element of it).

We define also the projection of a response equivalence class $R \in P_R$ on $P_K$ as

$$\sigma(R) = \{B \mid B = R \cap K_i \text{ with } K_i \in P_K\},$$

and the extended projection

$$\sigma^*(R) = \{K_i \mid K_i \in P_K \text{ and } K_i \cap R \neq \emptyset\}.$$

The same inequalities hold for $\sigma(R)$:

$$1 \leq |\sigma(R)| \leq |R|.$$

---

[2] F-measure (F) is defined by $2/F = 1/R + 1/P$.

[3] Cf. the paragraph on "The Problem of RE Identification".

[4] Actually, the MUC method considers for K only MRs having two or more REs (i.e. at least one coreference); yet, it can be shown that considering also singletons doesn't change the results (cf. note 3).

[5] |X| stands for the cardinal number of the set X, i.e. the number of elements in X.

**Computer Interface** Understanding of the projections is made clearer by our graphical interface (Table 1). Both the left and the right columns display the elements of *R*, structured according to $P_K$ (left) and $P_R$ (right): one equivalence class per line. Given the key and response partitions of the RE set (using the SGML/MUC format or directly Smalltalk objects), the interface displays on demand $\pi^*(K_i)$ and $\sigma^*(R_j)$, when the user double-clicks on a $K_i$ or a $R_j$ line.

The toy example shown below has 17 REs, noted 1, 2,…, 17; they correspond to four key discourse referents, $K_1$, $K_2$, $K_3$, $K_4$, shown on four separate lines. The system's (fictitious) response has three discourse referents, $R_1$, $R_2$, $R_3$ and quite a different grouping of the REs. A double-click on $K_3$ highlights $K_3$ itself, as well as its projections, $R_1$ and $R_2$. Conversely, a double-click on $R_1$ highlights $K_1$ and $K_3$. This interface is of course very useful to inspect the system's response during development.

| Key | Response |
|---|---|
| $K_1$ : 1, 2 | **$R_1$ : 1, 2, 6, 7, 8, 9, 10** |
| $K_2$ : 3, 4, 5 | **$R_2$ : 3, 4, 5, 11, 12, 13, 14, 15, 16** |
| **$K_3$ : 6, 7, 8, 9, 10, 11, 12** | $R_3$ : 17 |
| $K_4$ : 13, 14, 15, 16, 17 | |

Table 1: Principle of the interface for visual evaluation

### The Problem of RE Identification

One of the difficulties of the MUC-6 coreference task was that referring expressions had to be recognized by the system itself, which isn't strictly speaking part of the coreference task. Strict evaluation of reference resolution should probably be made on fixed RE sets, but, of course, a complete system is able to build them as well.

Here, we describe algorithms as if key and response partitions were indeed partitions of the same set. In fact, the MUC algorithm as well as ours operates also when these aren't partitions of the same set. Our formalism introduces then the notion of *extension* of an RE set: both the key and the response set are extended using the missing elements so that their extensions be the same. In this way we are brought back to the current formalism; of course the interesting point is the proof that extensions do not really change the formulae proposed here. As full detail cannot be given here, we keep analyzing the simple case.

### The MUC Algorithm

In order to clarify the use of our frame and relate our work to the MUC scoring method, we give briefly the corresponding formulae. We use the following notation: MRE for MUC recall error, MPS for MUC precision success (score) and so on.

**Description** For a given key equivalence class K, the recall error is the number of missing links between the different projections of K on $P_R$. According to Vilain et al. (1995), it would be too severe to count the missing links between all the REs of all projections of K: an indulgent evaluator counts only the minimal number of coreferences needed to link all the projections of K. Thus, MRE(K) = $|\pi(K)| - 1$ ; MRE(K) varies between 0 (K isn't split at all in the response) and $|K| - 1$ (K is completely split), so MRE(K) is normalized by $|K| - 1$. When summing on all K in $P_K$, and given that MRS = 1 − MRE, we obtain:

$$MRS = ( |R| - \Sigma_K |\pi(K)| ) / ( |R| - |P_K| ).$$

An elegant feature of the MUC algorithm is that precision is defined symmetrically, because a response equivalence class $R_j$ having two projections on $P_K$ means that the system has wrongly aggregated two referents which are distinct in reality, so there is one wrong extra link (indulgent evaluation). So,

$$MPS = ( |R| - \Sigma_R |\sigma(R)| ) / ( |R| - |P_R| ).$$

It can be shown that in the MRS and MPS formulae, numerators are always equal: this coincidence is visible on the MUC-6 scoring reports, p.316. It can be proved using the partition of *R* formed by all the intersections of each $K_i$ and $R_j$, partition whose cardinal number equals both $\Sigma_K |\pi(K)|$ and its symmetric $\Sigma_R |\sigma(R)|$.

**Shortcomings** This algorithm is maximally indulgent: it computes the minimal number of errors that may be attributed to the program. But in some cases this isn't a relevant figure. For instance, on texts with high coreference rates, massive (and meaningless) overgroupings of REs by a program aren't much penalized because they can be indulgently attributed to only a very few excedentary coreferences (see our example Table 2, line (2), and more numeric results below).

## Evaluation Using Core Equivalence Classes

To find a more relevant evaluation method, one should keep closer to the idea of discourse referents. Such a method should first be able to find out which response equivalence class $R_j$ among the classes of the projection $\pi^*(K_i)$ represents the program's representation of a given key referent $K_i$; this $R_j$ will be called *the core MR of the referent $K_i$*. The method should afterwards count each RE which is not attached to the core $R_j$ as a recall mistake for $K_i$ each RE, and not each group of REs, as in the MUC method. Our first algorithm computes precision symmetrically. The second algorithm adds disjunction constraints on core MRs and computes precision asymmetrically, using *exclusive core MRs*.

### First Algorithm: Core MRs

**Definition** For a given discourse referent K, its core MR is the R∈$P_R$ containing the greatest projection of K, i.e. the greatest number of REs from K. Let c(K) be the greatest projection of K,

$$c(K) = \max(A) \text{ for } A \in \pi(K).$$

The core MR of K noted c*(K) is the R which contains c(K),

$$c^*(K) = R \text{ where } R \supset c(K), R \in P_R.$$

In order to compute precision, we define the symmetric core MR for a given R∈$P_R$, noted c*(R)∈$P_K$.

**Scoring Method** All the REs belonging to a given K but not to its core MR c(K) are counted as recall errors, regardless of whether they are grouped together or not. So, CRE(K) (the core recall error for K) is

$$CRE(K) = \sum_{\substack{K_i \in \pi^*(K) \\ K_i \bullet c(K)}} |K_i| = |K| - |c(K)|.$$

This varies between 0 (all REs from K are correctly grouped together in a single response MR) and |K| – 1 (all REs are separated, K is completely split), so CRE(K) has to be normalized by |K| – 1.
As CRS = 1 – CRE, when summing on all the key equivalence classes K we obtain the following core recall success (or score):

$$CRS = \left( \sum_{K \in P_K} |c(K)| - |P_K| \right) / (|R| - |P_K|) \ [6].$$

Precision is computed symmetrically using the key core classes for each response MR, namely $c(R_j)$ or $c^*(R_j)$. The core precision score (or success) is:

$$CPS = \left( \sum_{R \in P_R} |c(R)| - |P_R| \right) / (|R| - |P_R|).$$

Before looking at some examples, we give a formal comparison between this method and the MUC one.

**Theoretical Comparison** The core-MR scoring algorithm is less indulgent than the MUC one, as individual REs contribute each to the error score. To show that CRS • MRS we will show that for all K∈$P_K$, CRS(K) • MRS(K). By definition and rearrangement,

$$CRS(K) \bullet MRS(K) \Leftrightarrow |\pi(K)| - 1 \bullet |K| - |c(K)|.$$

The second inequality states precisely that the number of projections of K except the core c(K) itself is smaller than the number of elements of K which are not in the core, which is clearly true because each projection has at least one element. The equality happens if and only if all projections of K on $P_R$, except the core, are singletons.
For instance, if K = {1, 2, 3, 4, 5, 6} and π(K) = {{1, 2, 3}, {4}, {5}, {6}}, then MRS(K) = CRS(K) = 2/5. But if π(K) = {{1, 2, 3}, {4, 5, 6}}, then MRS(K) = 4/5 and CRS(K) = 2/5. In terms of coreference links, the core-MR method estimates in the second case that only two links out of five have been found (say, 1-2 and 1-3, but not 1-4, 1-5 and 1-6) whereas the MUC method counts only one missing link (say, 3-4). On longer texts with high coreference rates, the MUC algorithm provides counterintuitively high success scores even for simplistic solving methods, as shown below.

**Numeric Comparison** Comparative results on two short examples are given Table 2 below. The first is a fictitious text with ten referring expressions, and two MRs, $K_1$ = {1, 2, 3, 4, 5} and $K_2$ = {6, 7, 8, 9, 10}. We suppose first (1) that the system has done no resolution, that is $R_1$ = {1}, $R_2$ = {2}, ..., $R_{10}$ = {10}. Both scoring methods reasonably return 0% recall and 100% precision. But let us suppose (2) that the evaluated system simply groups all REs together, $R_1$ = {1, 2, ..., 10}. Then MPS = 89% which is a very high figure for a very poor method, while CPS = 44% seems more relevant.

The second text is the Walkthrough article used for MUC-6 (Sundheim & Grishman 1995, p. 283). This text has 147 referring strings intervening in coreference relations[7]. These expressions are grouped in 15 key equivalence classes or MRs ; there are 50 pronouns, but only 5 MRs contain pronouns.
Let us suppose (3) that the pronouns aren't solved at all by a system, but all the other REs are perfectly identified and solved: MRS = CRS = 62% is a reasonable figure. But suppose now (4) that the same system groups together all the pronouns in a 16th class (this operation needs hardly any knowledge): it obtains 96% success from the MUC algorithm, which is clearly exaggerated. Let us imagine finally (5) a dumb system which groups all 147 REs in a single response MR; the MUC algorithm credits it with 90% precision[8]. Recall being still 100%, this is far better than any of the competing systems in MUC-6. Of course, those systems had to identify previously the correct REs, which is a difficult and penalizing task.

| Text | MRS | MPS | CRS | CPS |
|---|---|---|---|---|
| (1) 10 REs / no resolution | 0 | 100 | 0 | 100 |
| (2) 10 REs / 1 MR | 100 | 89 | 100 | 44 |
| (3) W / no pron. resolution | 62 | 100 | 62 | 100 |
| (4) W / grouped pron. | 97 | 96 | 63 | 81 |
| (5) W / 1 MR | 100 | 90 | 100 | 31 |

Table 2: MUC and core-MR scores (%) for different keys and responses

**Second Algorithm: Exclusive Core-MRs**

Doubts may be cast on the cognitive relevance of the core-MR method when noticing that several key MRs may share the same core MR. This will naturally count a lot for precision errors, but may seem strange if core MRs are conceived as the skeleton of the program's understanding of referents. It seems more sound to suppose that core MRs have to be distinct; thus, the evaluation method should first find out in the response the exclusive core MRs using the algorithm we give below, and then try to match these MRs against the key MRs and compute recall and precision.
The *exclusive core MR* for a given key MR $K_i$, noted $xc^*(K_i)$, is "the image the system has of the referent $K_i$". To remain close to this view, there is no reciprocal notion of exclusive core for a response MR: precision is computed using the same $K_i$ and $xc^*(K_i)$, not using a symmetric construction.

**Definition** The main problem with the function $c^* : P_K \to P_R$ is that it is not an injective function (several elements K of $P_K$ can have the same core MR). In order to associate an unique xc(K) to each K, we start with the K∈$P_K$ having the greatest projection π(K), and define xc(K) = c(K) and $xc^*(K) = c^*(K)$. Then, we remove K from $P_K$ and xc*(K) from $P_R$, and we look for the next K having the greatest projection on $P_R' = P_R - \{xc(K)\}$, and associate it xc(K); and so on. At the end, there might be some K∈$P_K$ which have no projection left: then, xc(K) = ∅, and these K are viewed as completely unrecognized referents.

---

[6] If |$P_K$| = |R|, i.e. all the key MRs are completely split, there is no coreference to be solved, the result is zero despite the zero denominator, because all the CRS(K) are null.

[7] There are actually more REs, but only the ones intervening in coreference relations matter, as shown previously.
[8] This figure comes from the fact that R being the only response equivalence class, MPS = (|R|–|σ(R)|) / (|R|–1) = (147–15) / (147–1) = 0.904 .

**Scoring Method** After all xc(K) are determined, recall and precision are computed in a simple way: all the REs not attached to an exclusive core MR are recall errors, and REs attached to the wrong exclusive core MR are precision errors. So, the exclusive core recall score is:

$$XRS = ( \sum_{K \in P_K} |xc(K)| ) / |R|, \text{ and}$$

$$XPS = 1 - XPE = 1 - ( \sum_{K \in P_K} |xc^*(K) - K| ) / |R|.$$

**Variations** The limits of variation of XRS and XPS are less obvious than the other ones. For a given K, XRS(K) = |xc(K)|/|K| varies between 0 (no exclusive core MR for K) and 1 (all the REs of K are grouped into xc(K)), but as there is always at least one exclusive core for $P_K$, the total XRS cannot be zero. It can only reach 1/|R|, which can be arbitrarily small. Total XPS varies from 1/|R| to 1 (when no exclusive core MR contains any foreign RE).[9]

**Example** Evaluation (Table 3) of our toy example (Table 1) makes the method clearer. To assign exclusive core MRs, the algorithm starts with K3, because |c(K3)| = 5. So xc*(K3) = R1. Then xc*(K4) = R2, xc*(K2) = ∅, xc*(K3) = ∅. Scores are given in Table 3 for the three algorithms: the last method appears to be the most severe, followed by the core-MR method.

|  | MRS | MPS | CRS | CPS | XRS | XPS |
|---|---|---|---|---|---|---|
| Fraction | 11/13 | 11/14 | 10/13 | 7/14 | 9/17 | 9/17 |
| Value | .85 | .79 | .77 | .50 | .53 | .53 |

Table 3: Scores for the example given Table 1

## The Distributional Algorithm

We finally outline a tentative scoring method aimed not at the exact count of errors for each MR, but at characterizing the aspect of the global distribution of REs into MRs. It is therefore not an exact measure of reference resolution, but one giving a global view of the main trend of a system.

### Description
**Distribution Overlapping** The distribution of REs into MRs is the list of the sizes of each MR, sorted by decreasing size. For the example Table 1, the key distribution is (7, 5, 3, 1) and the response one (9, 7, 1). A first measure of their similitude is their overlapping, i.e., the number of different REs in the first MR plus the second plus the third, etc. If one of the distributions runs out of MRs for this comparison (here, the response lacks a fourth element), null MRs are added at the end. Overlapping is thus 7+5+1+0 = 13 divided by the maximum value which is here (7+5+3+1) = (9+7+1) =17, so 76%.

---
[9] If R = {1, 2, ..., n}, K = {{1, 2, ..., n}} and R = {1}, {2}, ..., {n}}, then XRS = 1/n. If K and R are the other way round, then XPS = 1/n.

**D-error** It is still more interesting to find out whether the REs are aggregated more in the response than in the key, or less – in other terms whether the system assumes more coreferences than needed, or less, regardless of their intrinsic accuracy. Here, the biggest MR in the response is greater than the one in the key (9 vs. 7 REs), and the second too (7 vs. 5), but not the third. One way to measure this is to find out the average position (order) of the MRs for which the key MRs are greater than the response MRs, and the average position where the inverse holds. If the first average position is lower than the second (i.e. closer to the first and greatest MR), the system makes mainly *d-recall errors*: the greater response MRs have a lower size than the greater key MRs. In the opposite situation, when the greater response MRs are bigger than the key ones, the system finds too many coreferences, so it makes mostly *d-precision errors*. On the toy example, the d-error is 40%–precision. D-error is thus a measure of the distance between these two average positions, with a label "recall" or "precision" depending on their order. We won't give formulae for the distributional measures because of their complexity. However, we formalized these measures as well, and implemented them in our evaluation software.

### Analysis
Obviously, the distributional scores do not suffice to evaluate reference resolution: if the key and response distributions are the same, nothing guarantees that the corresponding MRs have indeed the same contents. It is however highly improbable that the response distribution match by chance the key one. Also, the other way round, if the key and response distributions don't match, we are certain that the response is different from the key.
The main interest of this method is to evaluate rapidly very deviant responses with a simple binary criterion ("too many coreferences" / "not enough coreferences"). Such deviant responses sometimes obtain good scores with the MUC scorer, as on line (5) in Table 2 (90% precision), while the distributional scoring finds only an overlapping of 31% and a d-error of 25%–precision, which shows that the system overgroups REs. More results are given below.

## Discussion

### Results on Our System
**The System** We are currently developing a reference resolution system implementing the MR-paradigm (Popescu-Belis & Robba 1997, 1998; Popescu-Belis 1998). The main ideas of the method are to use salience values for each MR, and to apply selectional constraints between the RE being solved and each MR, by averaging compatibility over all the REs of the MR. The system has yet little knowledge, using only a shallow parser (for NPs) and a small synonym dictionary.
One of the difficulties we encountered was performance assessment during development. We thus studied and implemented the MUC-6 algorithm, and were surprised by the system's good scores despite gross resolution mistakes. That is the reason why we developed more severe and relevant evaluation techniques, and implemented them along with the MUC-6 method.

**The Texts** The system was tested on two different texts: a short story by Stendhal (VA) and the first chapter of a novel by Balzac (LPG.long) – both 19th century French authors. Referring expressions were annotated manually, as well as coreferences between them[10]. The first text has 638 REs and 371 MRs, the second 3812 REs and 494 MRs. Both have high coreference rates, which partly explain the system's too good results for the MUC scoring. The third text (LPG.equ) is an excerpt from LPG.long having roughly the same size as VA. Input and output formats for these texts as well as pre-processing are detailed in (Popescu-Belis 1998). For instance, MUC-styled coreference links are a possible output.

**Results** Scores for the three methods are given Table 4. The core-MR method is clearly more strict than the MUC method; however, for both of them, the better results on LPG than VA do not reflect a system's improvement, but are due to a higher coreference rate for LPG. The core-MR figures reflect nevertheless better our own judgment of the system's performances.

|          | MRS | MPS | CRS | CPS | XRS | XPS |
|----------|-----|-----|-----|-----|-----|-----|
| VA       | .64 | .57 | .41 | .37 | .56 | .69 |
| LPG equ1 | .75 | .76 | .61 | .41 | .41 | .53 |
| LPG long | .83 | .88 | .58 | .40 | .34 | .50 |

Table 4: Scores of our system for the three texts

While M_S and C_S in Table 4 follow globally the same pattern but at different levels, the X_S scores correctly show the opposite order. They actually prove closer to the distributional scores (see Table 5), which are the lowest on LPG.long, both for d-error and overlapping. So, high coreference rates reduce the MUC and core-RM relevance, but not that of the exclusive-core-RM and distributional algorithms.

|          | D-error              | Overlapping |
|----------|----------------------|-------------|
| VA       | 28,3% — precision    | 80,0%       |
| LPG equ1 | 32,8% — precision    | 63,2%       |
| LPG long | 36,7% — precision    | 51,4%       |

Table 5: D-scores of our system for the three texts

**Evaluating evaluation ?**

It is not very clear which arguments could prove that a scoring method is more valuable that another. We argued previously that an evaluation method had to capture the human evaluator's view of a good or poor program. However, judgments of human reference resolution are often limited to "right or wrong", and poor understanding (by children or foreigners) is often qualified as "no understanding at all".

In these conditions, we have outlined some minimal criteria for sound evaluation methods, as well-defined bounds and scalability (between 0 and 100%). We may add now that such methods have to be enough "linear": low scores have to be as probable as higher ones, or high scores shouldn't be too easily reached. Linearity assessment for a given method is a complex task, but it can be performed using simple toy examples as the one we used. This would constitute then a genuine way of "evaluating evaluation".

**Conclusion**

We have presented in this paper three new methods for reference resolution evaluation. Only their testing and use by different teams could help finding out their qualities and shortcomings. Our further work will focus on improvment of our resolution system, assessed stage by stage using all of the four scoring methods discussed here.

### Acknowledgments
This work is part of the Cervical Project supported by the GIS Sciences de la Cognition.

---

[10] At LORIA, Nancy, France, for the second one.